\documentclass[10pt,conference]{IEEEtran}
\pdfoutput=1
\usepackage[utf8]{inputenc}
\usepackage{times}
\usepackage{epsf}
\usepackage{threeparttable}
\usepackage{setspace}
\usepackage{amsmath}
\usepackage{amsthm}
\usepackage{epsfig}
\usepackage{caption}
\usepackage{subcaption}
\usepackage{graphicx}
\usepackage[square,numbers,sort]{natbib}
\usepackage{hyperref} % hyperlinks for references.
\usepackage{algorithm}
\usepackage{algpseudocode}
\usepackage{listings}
\usepackage{color}
\usepackage{color}
\usepackage{tcolorbox}
\usepackage{array}
\usepackage{booktabs}
\usepackage{adjustbox}
\usepackage{dirtytalk} %!!!
\usepackage[T1]{fontenc}
\usepackage{cleveref}

\definecolor{mygreen}{rgb}{0,0.6,0}
\definecolor{mygray}{rgb}{0.5,0.5,0.5}
\definecolor{mymauve}{rgb}{0.58,0,0.82}

\title{Accelerating Genetic Programming using GPUs}
\author{

    \IEEEauthorblockN{
        Vimarsh Sathia\IEEEauthorrefmark{1}, 
            Venkataramana Ganesh\IEEEauthorrefmark{2},
        Shankara Rao Thejaswi Nanditale\IEEEauthorrefmark{2}
    }

    \IEEEauthorblockN{cs17b046@cse.iitm.ac.in, gvenkatarama@nvidia.com, snanditale@nvidia.com}

    \IEEEauthorblockA{\IEEEauthorrefmark{1}\textit{Department of CSE}, \textit{IIT Madras}, \textit{India} \\}
    \IEEEauthorblockA{\IEEEauthorrefmark{2}\textit{NVIDIA Corporation}}
}

\begin{document}
\maketitle

\begin{abstract}
Genetic Programming (GP), an evolutionary learning technique, has multiple applications in machine learning such as curve fitting, data modelling, feature selection, classification etc. GP has several inherent parallel steps, making it an ideal candidate for GPU based parallelization. This paper describes a GPU accelerated stack-based variant of the generational GP algorithm which can be used for symbolic regression and binary classification. The selection and evaluation steps of the generational GP algorithm are parallelized using CUDA. We introduce representing candidate solution expressions as prefix lists, which enables evaluation using a fixed-length stack in GPU memory. CUDA based matrix vector operations are also used for computation of the fitness of population programs. We evaluate our algorithm on synthetic datasets for the Pagie Polynomial (ranging in size from $4096$ to $16$ million points), profiling training times of our algorithm with other standard symbolic regression libraries viz. \texttt{gplearn}, \texttt{TensorGP}, \texttt{KarooGP}. In addition, using $6$ large-scale regression and classification datasets usually used for comparing gradient boosting algorithms, we run performance benchmarks on our algorithm and \texttt{gplearn}, profiling the training time, test accuracy, and loss. On an NVIDIA DGX-A100 GPU, our algorithm outperforms all the previously listed frameworks, and in particular, achieves average speedups of $119\times$ and $40\times$  against \texttt{gplearn} on the synthetic and large scale datasets respectively.
\end{abstract}
\section{Introduction}
\label{sec:intro}
In recent years, there has been a widespread increase in the use of GPUs in the field of machine learning and deep learning, because of their ability to massively speedup the training of models. This is mainly due to the large parallel processing ability of GPUs, which can process multiple inputs using Single Instruction Multiple Data (SIMD) intrinsics. Genetic Programming (GP) belongs to a class of machine learning algorithms with several inherent parallel steps. As such, it is an ideal domain for GPU based parallelization. 

GP as a technique involves the evolution of a set of programs based on the principles of genetics and natural selection. It is a generalized heuristic search technique, which searches for the best program optimizing a given fitness function. Because of it's generalizability, the technique finds applications ranging from machine learning to code synthesis \citep{Koza92}. %It also supports a meta-evolutionary framework, where a GP system itself can be evolved using GP\citep{schaul2010metalearning}.

However, in practice, it is difficult to scale GP algorithms. Fitness evaluation of candidate programs in GP algorithms is a well-known bottleneck, and there are multiple previous attempts to overcome this problem - either through parallelization of the evaluation step
% \citep{10.1007/978-3-540-71605-1_9,baeta2021tensorgp,DEAP_JMLR2012,gplearn,staats2017tensorflow}
, or by eliminating the need for fitness computations itself through careful initialization and controlled evolution \citep{biles2001autonomous}.

In this paper, we introduce a parallelized version of the generational GP algorithm to solve symbolic regression problems. We use a stack-based GP model for program evaluation (inspired by \citep{perkis}). Our algorithm can also be used to train binary classifiers, where the classifier output corresponds to the estimated equation of the decision boundary. 

We evaluate our accelerated algorithm by providing execution time benchmarks for program evaluation. The same benchmarks are also run on other standard libraries like gplearn \citep{gplearn} and TensorGP \citep{baeta2021tensorgp}, for comparing training speeds. We also study the effect of population size and dataset size on the final time taken for training. The benchmarks are run using synthetic datasets for the Pagie Polynomial \citep{Pagie1997} as well as $6$ large-scale regression and classification datasets usually used for the comparison of gradient boosting frameworks \citep{Dua:2019}.  

Our work has been implemented as an algorithm in \texttt{cuML} \citep{raschka2020machine}, a GPU accelerated machine learning library with an API inspired by the \texttt{scikit-learn} library \citep{scikit-learn}. 

The rest of this paper is organized as follows. \Cref{chap:bgrw} gives an overview of existing literature, and introduces the generational GP algorithm. It also discusses other existing GP frameworks and libraries. \Cref{chap:ourwork} presents our parallel algorithm to perform symbolic regression using genetic programming, along with some implementation details and challenges faced. \Cref{chap:experiments} describes our experimental setup and presents our benchmarking results. Finally \Cref{chap:conclusion} concludes the paper and outlines directions for further optimizations and future research. 

\section{Background and Related Work}
\label{chap:bgrw}
In this section, we describe in detail the generational GP algorithm \citep{poli08:fieldguide,RFIgp2016}, which serves as a base for all of our parallelization experiments. Later on, we also give a brief description of some existing GP libraries. 
 
Since the domain of our problem is symbolic regression, we note that the program for our GP system is a mathematical expression, represented as an expression tree. Each program comprises a list of functions (of varying arity) and terminals, where terminals collectively denote both variables and constants.
\subsection{The Generational GP Algorithm}
\label{bgrw:algo}
The generational GP algorithm gives us a method to evolve candidate programs using the principles of natural selection. There are $3$ main steps involved in the algorithm. 
\begin{enumerate}
  \item Selection --- In this step, we decide on a set of programs to evolve into the next generation, using a selection criterion. 
  \item Mutation --- Before promoting the programs selected in the previous step to the next generation, we perform some genetic operations on them. 
  \item Evaluation --- We again evaluate the mutated programs on the input dataset to recompute fitness scores. 
\end{enumerate}

The above $3$ steps are formalized in \Cref{gpalgo}.
% \Cref{gpalgo} formalizes the above $3$ steps. The initial set of programs is usually randomly generated. The termination criterion defined in \Cref*{gpalgoendloop} of the algorithm is usually activated when the number of generations reaches a maximum limit or an early convergence is reached on the given input dataset (usually governed by a user-specified threshold on fitness).

% In \Cref{subsec:selection} to \Cref{subsec:evaluation}, we dive deep into the selection, mutation and execution steps of the generational GP algorithm. In \Cref{subsec:mutation}, we define and implement $4$ different possible types of mutation along with reproduction.

\begin{algorithm}
  \caption{The Generational GP Algorithm}\label{gpalgo}
  \begin{algorithmic}[1]
  \Procedure{GP-FIT}{$dataset$}
  \State $curr \gets $Initialize population 
  \State \Call{Evaluate}{$curr, dataset$} 
  % \State 
  \Repeat
  \State $next \gets $ \Call{Select}{$curr$}
  \State $next \gets $ \Call{Mutate}{$next$} 
  \State \Call{Evaluate}{$next,dataset$}
  \State $curr \gets next$
  \Until{user defined termination criteria on $curr$ not met}\label{gpalgoendloop}
  \State \textbf{return} $curr$\Comment{The final generation of programs}
  \EndProcedure
  \end{algorithmic}
\end{algorithm}

% Before going into selection, we first talk about initialization methods for the first generation programs. 
Koza \citep{Koza92} lists $3$ standard initialization techniques for GP programs. They are as follows.
\begin{itemize}
    \item Full initialization --- All trees in the current generation are \say{dense}, that is, all the terminals(variables or constants), are at a distance of \emph{max\_depth} from the root.
    \item Grow initialization --- All trees grown in the current generation need not be \say{dense}, and some nodes that have not reached \emph{max\_depth} can also be terminals. 
    \item Ramped half-and-half --- Half of the population trees are initialized using the Full method, and the other half is initialized using the Grow method. (the common usage)
\end{itemize}

% In addition to the above $3$ methods, we note that there are other initialization algorithms which grant the user more control over tree initialization with respect to tree depth and node probabilities\citep{luke:2000:2ftcaGP}. However, we provide support only for the $3$ standard initialization methods in our implementation.

\subsubsection{Selection}
\label{subsec:selection}
Selection is the step where individual candidates from a given population are chosen for evolution into the next generation. According to Goldberg and Deb \citep{GOLDBERG199169}, the commonly used selection schemes in GP are as follows:
\begin{itemize}
  \item Tournament selection --- Winning programs are determined by selecting the best programs from a subset of the whole population(a \say{tournament}). Multiple tournaments are held until we have enough programs selected for the next generation. 
  \item Proportionate selection --- Probability of candidate program being selected for evolution is directly proportional to its fitness value in the previous generation
  \item Ranking selection --- The population is first ranked according to fitness values, and then a proportionate selection is performed according to the imputed ranks. 
  \item Genitor(or \say{steady state}) selection --- The programs with high fitness are carried forward into the next generation. However, the programs with low fitness are replaced with mutated versions of the ones with higher fitness.
\end{itemize}

In the implementation of our algorithm, only parallel tournament selections are supported through the use of a CUDA kernel. This is in an effort to make our implementation consistent with \texttt{gplearn} \citep{gplearn}, which also only implements tournament selection. 

\subsubsection{Mutation}
\label{subsec:mutation}
As discussed earlier, the winning programs after selection are not directly carried forward into the next generation. Rather, \textit{mutations} or genetic operations are applied on the selected programs, in order to produce new offspring. Stephens \citep{gplearn} lists some commonly used mutation operations, which are as follows.
\begin{itemize}
  \item Reproduction
  \item Point mutations
  \item Hoist mutations
  \item Subtree mutations
  \item Crossover mutations
\end{itemize}

In our code we provide support for all the above listed genetic operations, along with a modified version of the crossover operation to account for tree depth.

In \emph{Reproduction}, as the name suggests, the current winning program is simply cloned into the next generation of programs.

% \begin{figure}
%   \centering
%   \begin{subfigure}{\columnwidth}
%     \begin{adjustbox}{width=\columnwidth,center}
%       \includegraphics{images/graphviz/point_mut_before.dot.pdf}
%     \end{adjustbox}
%     \caption{The original expression tree before performing a point mutation}
%     \label{fig:point_muta}
%   \end{subfigure}
%   \\
%   \begin{subfigure}{\columnwidth}
%     \begin{adjustbox}{width=\columnwidth,center}
%       \includegraphics{images/graphviz/point_mut_after.dot.pdf}
%     \end{adjustbox}
%     \caption{The same expression tree after a point mutation. The shaded nodes represent the changed terminals and functions in the final program.}
%     \label{fig:point_mutb}
%   \end{subfigure}
%   \caption{Visualizing point mutations for a given program.}
%   \label{fig:point}
% \end{figure}

In \emph{Point mutations}, we modify randomly chosen nodes of a given parent program in place. In the context of symbolic regression, we replace terminals(variables or constants) with terminals, and functions with another functions of the same arity. This mutation has the effect of reintroducing extinct functions and variables into the population, helping maintain diversity. %\Cref{fig:point} visualizes the effect of point mutations on an example program. % given program, where a subset of the functions and terminals in the original program in \Cref{fig:point_muta} are replaced with new functions and terminals in \Cref{fig:point_mutb} respectively.

% \begin{figure}
%   \centering
%   \begin{subfigure}{\columnwidth}
%     \begin{adjustbox}{width=\columnwidth,center}
%       \includegraphics{images/graphviz/hoist_mut_before.dot.pdf}
%     \end{adjustbox}
%     \caption{The original expression tree before performing a hoist mutation. The dark grey nodes denote the selected subtree, and the grey nodes denote the hoist subtree.}
%     \label{fig:hoist_muta}
%   \end{subfigure}%
  
%   \begin{subfigure}{\columnwidth}
%     \begin{adjustbox}{width=\columnwidth,center}
%       \includegraphics{images/graphviz/hoist_mut_after.dot.pdf}
%     \end{adjustbox}
%     \caption{The same expression tree after hoisting the sub-subtree onto it's parent's position.}
%     \label{fig:hoist_mutb}
%   \end{subfigure}
%   \caption{Visualizing hoist mutations for an expression tree.}
  
%   \label{fig:hoist}
% \end{figure}

In \emph{Hoist mutation}, a random subtree from the winner of the tournament is taken. Another random subtree is selected from this subtree as a replacement for the first subtree from the program. This mutation serves to reduce bloating of programs(with respect to depth) with increase in number of generations. %\Cref{fig:hoist} visualizes the effect of hoist mutations on a given expression tree. %In \Cref{fig:hoist_muta}, the left subtree of the root node, and a subtree of the left subtree, shaded grey and light grey respectively, are chosen for hoisting. The entire left subtree is then replaced with just the chosen sub-subtree, as seen in \Cref{fig:hoist_mutb}.

\begin{figure}
  \centering
    \begin{subfigure}{\columnwidth}
      \begin{adjustbox}{width=\columnwidth,center}
        \includegraphics{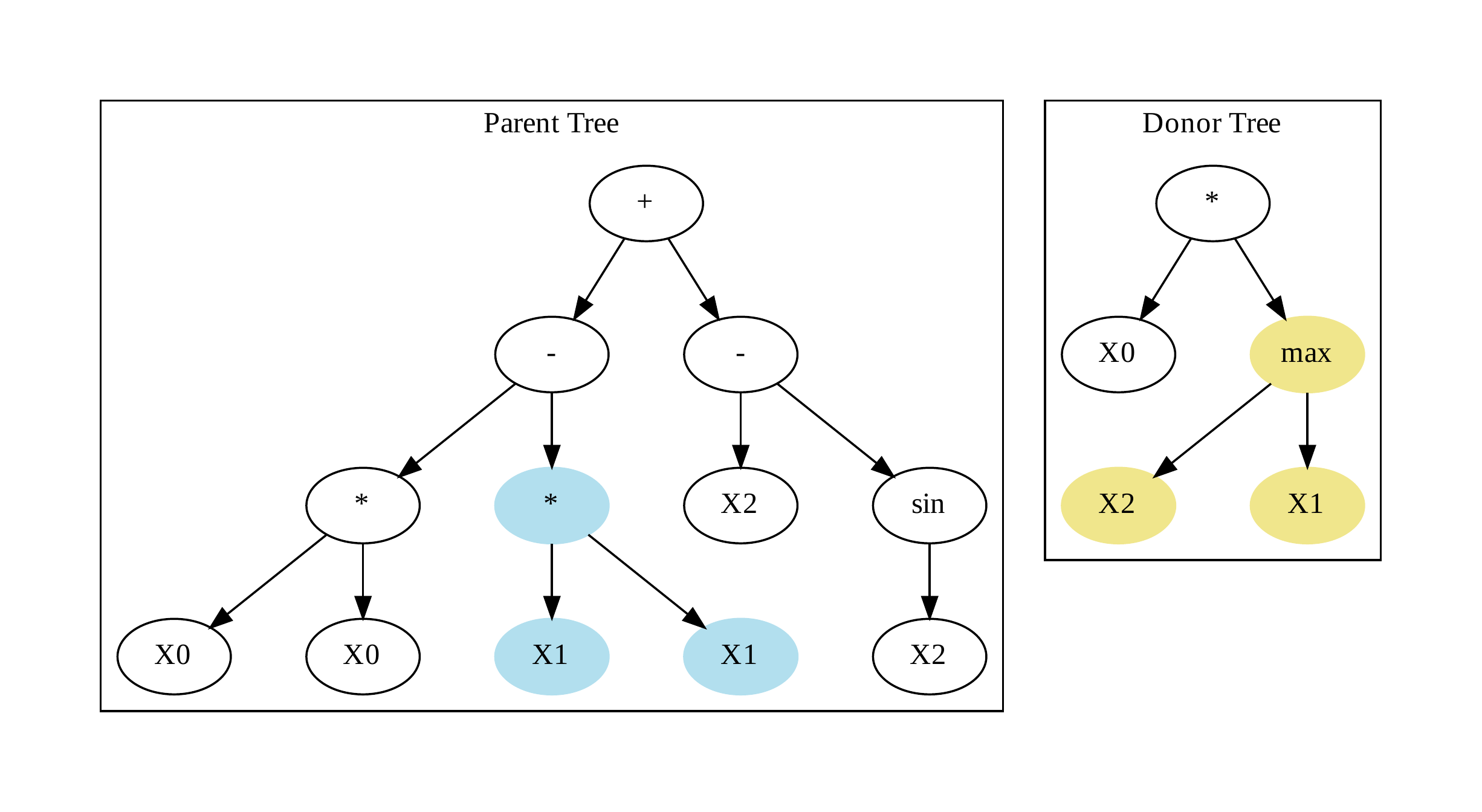}
      \end{adjustbox}
      \caption{The parent and donor expression trees, both selected through tournaments are shown here.}
      \label{fig:crossover_muta}
    \end{subfigure}%
    \\
    \begin{subfigure}{\columnwidth}
      \begin{adjustbox}{width=\columnwidth,center}
        \includegraphics{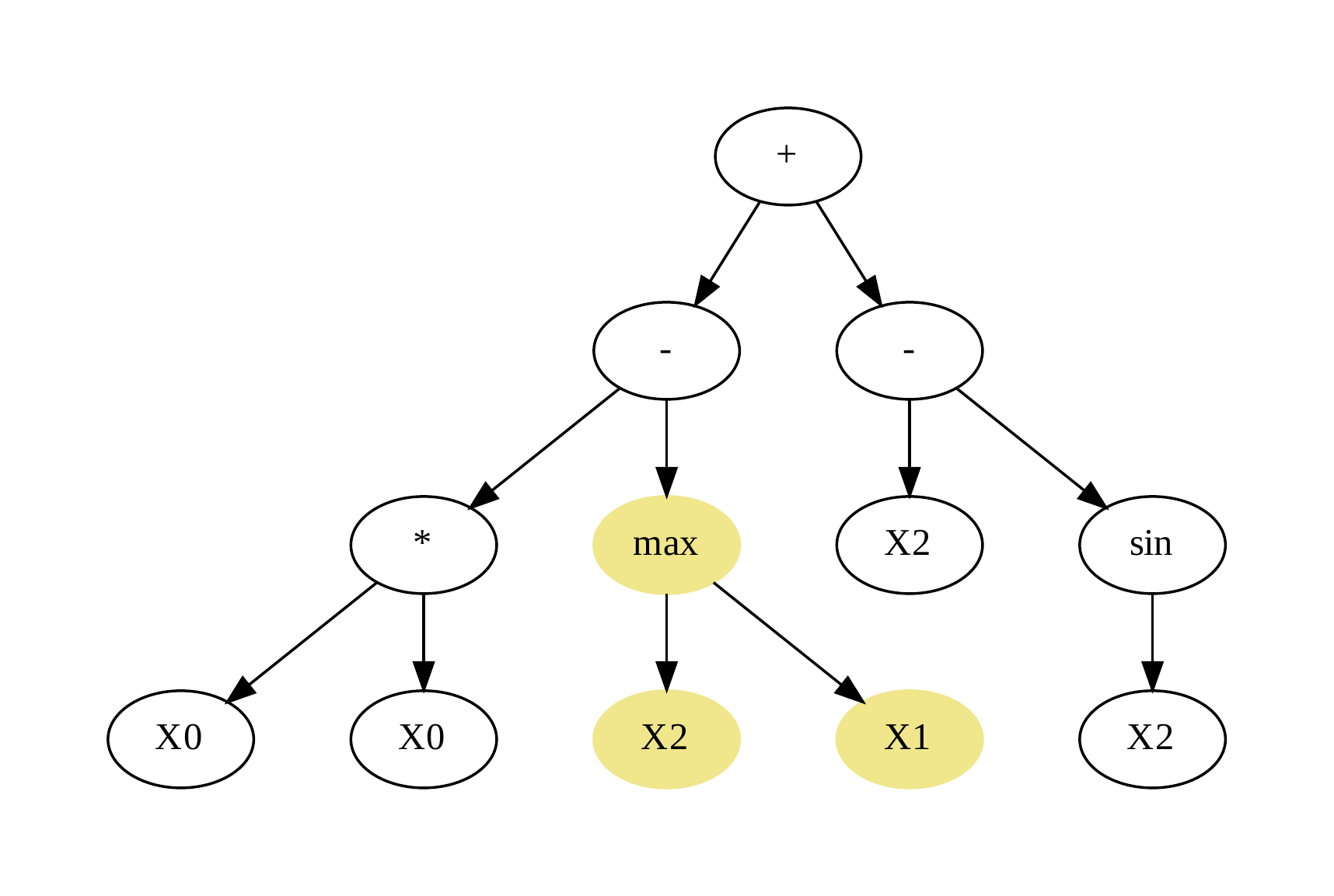}
      \end{adjustbox}
      \caption{The child expression tree after replacing a subtree of the parent with that of the donor.}
      \label{fig:crossover_mutb}
    \end{subfigure}
  \caption{Visualizing crossover mutations for a given parent and donor expression tree.}
  \label{fig:crossover}
\end{figure}

In \emph{Subtree mutation}, we perform a crossover operation between the parent tree and a randomly generated program. Similar to point mutations, this also serves as a method to increase new terminals and extinct functions in the next generation of programs.

In \emph{Crossover mutation}, the genetic information in $2$ programs of a given population are mixed. We first determine a parent and a donor program using $2$ separate tournaments. A random subtree of the parent program is then replaced with a subtree from the donor program. We show a sample visualization of the crossover operation in \Cref{fig:crossover}.

% Crossover is considered as the principal method for evolution in our implementation, which is again inspired by \citep{gplearn}.  %\Cref{fig:crossover_muta} contains both the parent and donor trees, along with random subtrees chosen from them. \Cref{fig:crossover_mutb} visualizes the nodes of the child tree after replacing the selected subtree in the parent with the donor subtree.

\Cref{fig:crossover} can again be used to visualize subtree mutations. The only difference between crossover and subtree mutations is that in subtree mutations, the donor program in \Cref{fig:crossover_muta} is now a randomly generated tree.

\subsubsection{Evaluation}
\label{subsec:evaluation}
Once the population for the next generation is decided after selection and mutation, the fitness of all programs in the new generation is recomputed. This step is the major bottleneck when scaling the generational GP algorithm to bigger datasets and bigger populations, as the evaluation is independent for every program and every row of the input dataset. 

% We perform the evaluation and computation of fitness in our implementation using a CUDA kernel and linear algebra primitives specified in the \texttt{raft} library\citep{raschka2020machine}.

\subsection{Existing Libraries and Their Coverage}
\label{sec:otherlibs}
\Cref{tab:otherlibs} (mostly reproduced from \cite{baeta2021speed}) lists some common libraries  used for genetic programming.  

\begin{table}[htbp]
  \caption{Some existing GP frameworks along with language and device support}
  \begin{center}
      \begin{tabular}[c]{ccc}
          \toprule
          \textbf{Framework} &   \textbf{Language} & \textbf{Compute Type} \\
          \midrule
          KarooGP & Python & CPU/GPU \\
          TensorGP & Python & CPU/GPU \\
          DEAP & Python & CPU \\
          gplearn  & Python & CPU \\
          ECJ & Java & CPU \\
          \bottomrule
      \end{tabular}
      \label{tab:otherlibs}
  \end{center}
\end{table}

Among the libraries listed in \Cref{tab:otherlibs}, both TensorG P\citep{baeta2021tensorgp} and KarooGP \citep{staats2017tensorflow} use the TensorFlow python framework to perform fitness evaluation the GPU. However, while KarooGP uses TensorFlow's \textit{graph} execution model, where the computation graph is compiled into an optimized DAG, TensorGP uses TensorFlow's \textit{eager} execution model, where expressions are evaluated immediately without the overhead of graph construction. The GPU parallelization here is through the use of TensorFlow-based vectorization, and not the explicit use of CUDA. 

DEAP \citep{DEAP_JMLR2012} is another commonly used Evolutionary Computing framework in Python implements a parallelized version of the GP framework. However, it offers only CPU based parallelization. 

Gplearn \citep{gplearn} is another Python framework which provides a method to build GP models for symbolic regression, classification and transformation using an API which is compatible with scikit-learn \citep{scikit-learn}. It also provides support for running the evolutionary process in parallel. The base code that is parallelized on GPUs in this paper is largely inspired by gplearn. 

The ECJ Evolutionary Computing Toolkit \citep{Luke1998ECJSoftware} is a Java library for many popular EC algorithms, with an emphasis towards genetic programming. Almost all aspects of \Cref{gpalgo} are governed by a hierarchy of user-provided parameter files and classes, and the framework itself is designed for large, heavyweight experimental needs. Using ECJ parameter files, it is also possible to define custom GP pipelines with user-defined evolution strategies.

\section{CUDA Accelerated Genetic Programming}
\label{chap:ourwork}
This section presents the implementation details of our parallel algorithm to perform genetic programming using CUDA. We first talk about a way to represent programs on the GPU. This is then followed by a description of the device side data structures used. We then give an overview of our modified GP algorithm, describing GPU-side optimizations for the selection, and evaluation step in detail. We also include details about the fitness computation step which comes after the evaluation step. Finally, we talk about the various challenges faced during the implementation of the modified algorithm, and the workarounds to avoid these problems.

In this implementation, we use a fixed list of functions with a maximum arity of $2$. 
% We assume that the maximum depth of all expression trees is fixed at $20$. 
We assume that the input dataset and actual output provided for training already resides in GPU memory, and are stored in a column-major order format.

\subsection{Program Representation}
\label{ow:input}
We define a struct for a program containing the following components. 
\begin{itemize}
  \item An array of operators and operands of the underlying expression tree stored in Polish (prefix) notation.
  \item A length parameter corresponding to the total number of nodes in the expression tree. 
  \item A raw fitness parameter containing the score of the current tree on the input data-set. 
  \item The depth of the current expression tree. 
\end{itemize}

\Cref{lst:programstruct} shows the internal definition of the program struct used in our code. \lstinline!nodes! is the prefix list used to store the nodes of the underlying expression tree. The content of \lstinline!nodes! is decided through the mutation step. Evaluation and fitness computations are used for the computation of the \lstinline!raw_fitness_! field.

\begin{lstlisting}[caption={A simplified version of our code for the \lstinline!program! struct, representing a single expression tree. This entire structure is copied over and evaluated on the GPU.},label={lst:programstruct}]
struct program {
  node *nodes; // underlying AST nodes
  int len; 
  int depth;
  float raw_fitness_; // fitness on input dataset
  metric_t metric; // loss function type
};
\end{lstlisting}

The entire population for a given generation is thus stored in an Array of Structures (AoS) format. The evaluation using a stack is almost similar to the way the Push3 system \citep{push3Stack} evaluates GP programs, with the sole exception being the reverse iteration due to the prefix notation chosen for the trees. 

Prefix notation was used for the representation of nodes to aid with the process of generating random programs, where we directly generate a valid prefix-list on the CPU.

\subsection{Device Side Data Structures}
\label{ow:deviceds}
In order to perform tournament selection and evaluation on the GPU, we use the following device side data-structures. 
\begin{itemize}
  \item Philox Random Number Generator (RNG) --- We use the Philox counter-based RNG \citep{Philox2011} implemented in \texttt{raft} \citep{raschka2020machine} to generate random global indices for tournament selection inside the selection kernel. 
  \item Fixed size device stack --- We define a fixed size stack using a custom class, implementing the \textit{push}, \textit{pop} methods as \lstinline!inline __host__ __device__! functions. To avoid global memory accesses and encourage register look-ups for internal stack slots, the push and pop operations are implemented using an unrolled loop over all the available slots. 
%   This action is possible because the maximum size of the stack is fixed at $20$. 
\end{itemize}

Our kernels for both selection and program execution have been written in a way to eliminate any need for synchronization or barriers.

% \subsection{Memory Footprint}
% \label{ow:memory}
% We examine the memory footprint of our implementation with respect to the number of programs stored in both CPU and GPU memory below. 
% \begin{itemize}
%   \item CPU --- Depending on a user-specified flag, we maintain a history of all the programs for all generations. If this is not desired, then we only store information about $2$ generations, the current and the next generation until the end.
%   \item GPU --- Only the device memory corresponding to the current and the next generations of programs is stored through the run of the algorithm. During the course of the algorithm, the $2$ memory locations are updated with new programs in a ping pong fashion.
% \end{itemize}

% In the next section, we present our CUDA-based parallel GP algorithm.

\subsection{The Parallel GP Algorithm}
\label{ow:paralgo}
In this section, we again outline the individual steps of the generational GP Algorithm described in \Cref{bgrw:algo}. However, each step contains details specific to our implementation. For our implementation, the selection and evaluation steps are performed on the GPU, whereas mutations are carried out on the CPU.

Before performing any of the standard steps, we decide on the type of mutation through which the next generation program is produced. This mutation type selection is governed by user defined probabilities for the various types of mutations. This step is important, as we need to determine the exact  number of selection tournaments to be run (as crossover mutations require $2$ tournaments to decide the parent and donor trees). 

Once the required number of tournaments has been decided, we move on to the selection step.

\subsubsection{Selection}
\label{ow:selection}
Tournament Selections are carried out in parallel using a CUDA kernel. For a given tournament size $k$, after computing the required number of tournaments, a kernel is launched, where each thread corresponds to a unique tournament. Each thread performs the following set of computations.
\begin{itemize}
    \item Generate $k$ random program indices using a Philox RNG. 
    \item Find the optimal index value among the $k$ indices with respect to fitness values (after accounting for both the criterion and the parsimony coefficient penalty). 
    \item Record the optimal index computed in the previous step as the current thread's winner.
\end{itemize}

Depending on the type of the loss function, the criterion can either favour smaller or larger fitness function values. The parsimony coefficient is another constant parameter used to control bloating of programs, by adding a penalty proportional to the length of candidate programs. 
For a given program $p$ with parsimony coefficient $c$ and a criterion \emph{favouring smaller fitness values}, \Cref{eq:parsimony,eq:fitness} compute the adjusted fitness values used for computing optimal indices.

\begin{align}
    p.penalty &= c~(p.len) \label{eq:parsimony} \\
    p.fitness &= p.raw\_fitness + p.penalty \label{eq:fitness}
\end{align}

\subsubsection{Mutation}
\label{ow:mutation}
The mutation of programs takes place on the CPU itself. Since every program has its own specific mutation, a GPU based implementation would lead to significant warp divergence, especially when identifying sub-trees in a program (since a different length sub-tree would be selected for each program during crossover, subtree or hoist mutations).

We implement all the mutations mentioned in \cref{bgrw:algo} with a slight modification to the crossover operation in order to constrain the depth of the output tree. We call this modification a hoisted crossover.  
In a hoisted crossover operation, we initially perform a crossover between the parent tree and the donor tree. The selected subtree of the donor is then repeatedly hoisted onto the parent tree until the depth of the resultant tree is less than the maximum evaluation stack size. 

% Note that the hoist mutation occurs only on the donor sub-tree, since that is the part of the tree which contributes to the depth violation. If this was not the case, then another sub-tree of the parent sub-tree would have contributed towards the maximum depth violation, making the parent program itself an invalid one. 

This modification is necessary for our code since a stack of fixed size $m$ can only evaluate a tree of depth $m-1$ (assuming the maximum function arity is $2$).

At the end of mutations, we allocate and transfer the newly created programs onto GPU memory, in order to evaluate them on the input data-set. In order to save on device memory, we also deallocate the GPU memory of the previous population trees. Some of the challenges we faced due to the nested nature of program representation during the \lstinline!cudaMemcpy! operations between host and device memory are listed in \Cref{sec:challenges}. 

\subsubsection{Evaluation}
\label{ow:evaluation}
We divide the evaluation portion into $2$ steps, an execution step and a fitness metric computation step. 

In the execution step, all programs in the new population are evaluated on the given data-set, to produce set of predicted values. If $n$ is the population size, and $m$ is the number of samples in the input dataset, then we launch an execution kernel with a 2D grid of dimension $(\left\lceil m/256\right\rceil ,n)$ with $256$ threads per block. Each thread has its own device side stack which evaluates a prefix list based program on a specific row.

We note here that to avoid thread divergence in the execution kernel, it is important to ensure that each thread block executes on different rows of the same program. More details about this can be found in \Cref{sec:challenges}.

% Once we have the predicted values for all programs, we compute the raw fitness value of the every program with respect to the expected output using the user-defined metric in the fitness metric computation step. We now explore finer details of both steps

In the fitness metric computation step, for every program, we compute fitness using a user-selected loss function. The inputs to the loss function are the program's output values (from the execution step) and the actual outputs.

Since the fitness computation is the same for all programs, we computed loss for all programs in parallel on the GPU. The computation was structured as a $2$ step operation as follows. 
\begin{itemize}
    \item A matrix vector operation was used to compute row-wise loss for the input dataset, with the columns of the predicted value matrix corresponding to population programs. The actual output vector is broadcasted along all columns of the predicted value matrix in this step.
    \item For every column of the loss matrix computed in the previous step, a weighted sum reduction operation is carried out to get a vector containing final raw fitness values for all programs. 
\end{itemize}

Both of the above steps were implemented using the linear algebra and statistics primitives present in the \texttt{raft} library \citep{raschka2020machine}. In our implementation, weighted versions of the following $6$ standard loss functions were implemented:

% \Cref{lst:mse} contains our code for computing weighted mean square error using \texttt{raft} primitives. Most of the linear algebra primitives in \texttt{raft} are designed in a way to take in \lstinline!__device__! lambda functions as inputs for either reduction or pre(post)-processing of data, as evidenced by \lstinline!raft::linalg::matrixVectorOp! call in Line 14 of \Cref{lst:mse}.
    
% We implement a weighted version of the following $6$ standard loss functions:

\begin{itemize}
    \item Mean Absolute Error (MAE)
    \item Mean Square Error (MSE)
    \item Root Mean Square Error (RMSE)
    \item Logistic Loss (binary loss only)
    \item Karl Pearson's Correlation Coefficient
    \item Spearman's Rank Correlation Coefficient
\end{itemize}
    
During the implementation of Spearman's Rank Correlation, we used the \texttt{thrust} library from Nvidia to generate ranks for the given values. The Karl Pearson's coefficient is then computed for the imputed ranks. 
% \Cref{lst:Spearman} shows a snippet from our code for the rank computations using \texttt{thrust}. 
    
The default fitness function is set as MAE, in an effort to be consistent with \texttt{gplearn} \citep{gplearn}.

\subsection{Challenges Faced}
\label{sec:challenges}
In this section, we briefly talk about some challenges faced during the implementation of our parallel GP algorithm.

\subsubsection{Thread divergence and Global Memory Access}
\label{prob:divergence}
In the evaluation step, during function computations in the execution phase, we check for equality of the current function with $33$ pre-defined functions, using \lstinline!if-else! conditions on the node type. 

Since CUDA executes statements using warps of $32$ threads in parallel, when it encounters an \lstinline!if-else! block inside a kernel which is triggered only for a subset of the warp, both the \lstinline!if! and \lstinline!else! blocks are executed by all threads. During the execution of the \lstinline!if! block, the threads which do not trigger the \lstinline!if! condition are masked, but still consume resources, with the same behaviour exhibited for the \lstinline!else! block. This increases the total execution time as both blocks are effectively processed by all warp threads. 

To avoid this behaviour in our code, we ensure that within every thread-block of the execution kernel, all threads execute the same program. This ensures that all threads in a warp will always take the same branch during node identification, and thus avoid divergence during a single stack evaluation.

In the implementation of \textit{push} and \textit{pop} operations for the device side stack, we avoid trying a possible dereference using global memory index(the current number of elements in the stack) by using an unrolled loop for stack memory access. This is again safe from thread divergence because we ensure that within a thread-block, all threads evaluate the same program. 

\subsubsection{Memory Transfers and Allocation}
\label{prob:memcpy}
Since used an AoS representation for the list of programs and each program has a nested pointer for the list of nodes in it, we are forced to perform at least $2$ \lstinline!cudaMemcpy! operations per program in a loop spanning the population size. One of the copy operations is for the list of program nodes, and the other copy is to capture the metadata about the nodes, and the other copy is for the program struct itself. 

Since all our computations are ordered on a single CUDA stream, transferring programs back and forth the device in a loop slows down the overall time for training. However, in our experiments, we observed that the dominant contributor to execution time was the evaluation step, and not memory transfers.

\section{Experimental Evaluation}
\label{chap:experiments}
In this section, we will describe the various experiments ran to evaluate the performance of our algorithm's implementation(henceforth referred to as \textit{cuml}) against the other GP libraries mentioned in \Cref{sec:otherlibs}. In particular, we consider the \textit{gplearn} \citep{gplearn}, \textit{KarooGP} \citep{staats2017tensorflow} (only GPU), and \textit{TensorGP} \citep{baeta2021tensorgp} (both CPU and GPU) libraries in our benchmarks. 

We run $2$ sets of benchmarks on both synthetic and large-scale datasets. For the synthetic dataset benchmarks, we follow the framework laid out in \citep{baeta2021speed} to compare evaluation times for all the above GP libraries on seven 2D regression datasets ranging in size from $4096$ to $16$ million points for an average over $10$ runs. 

For the large-scale benchmarks, we consider $6$ real world datasets commonly used for the comparison of gradient boosting frameworks. We perform a detailed comparison of the performance of both \textit{gplearn} and \textit{cuml} on these datasets, profiling training time for an average over $3$ runs. Since \textit{gplearn} was our reference for implementation, we used it for comparison with \textit{cuml} on large-scale benchmarks. In addition, we also compare the test accuracy and loss for both libraries initialized with the same hyper-parameters.

Before exploring the benchmarks on the synthetic datasets, we note that field of GP, especially for symbolic regression suffers from a lack of standardized benchmarks. This problem is explored by a few studies \citep{GP_Better_Benchmarks}, which attempt to quantify and list candidate GP problems. In our experiments, we follow the guidelines laid out in these studies.

All experiments were carried out in a compute cluster with the specifications listed in \Cref{tab:laptop}.

\begin{table}[h]
  \caption{Hardware and software setup for carrying out all experiments}
  \begin{center}
      \begin{tabular}[c]{cc}%{|>{\bf}c|c|}
        \toprule
        \textbf{Component} &   \textbf{Specification}           \\
        \midrule
        CPU model  & Intel(R) Xeon(R) Silver 4110 @ 2.10GHz     \\
        GPU model  & NVIDIA DGX-A100                            \\
        % Driver Version & 465.18                            \\
        CTK Version & 11.2                                    \\ 
        \# CPU cores & 16                                       \\
        RAM         & 64 GB                                     \\
        OS          & Ubuntu 20.04 LTS Server                   \\
        \bottomrule
      \end{tabular}
      \label{tab:laptop}
  \end{center}
\end{table}
% \footnote{CTK - Cuda ToolKit}
\subsection{Synthetic Benchmarks}
\label{sec:synthetic}
Our synthetic benchmarks were motivated by \citep{baeta2021speed}, and follow a similar flow for testing execution times between different libraries. We also compare the variation of best fitness values for \textit{cuml} with increasing dataset size.

\subsubsection{Setup}
\label{subsec:setup}
In all symbolic regression runs, we try to approximate the Pagie Polynomial \citep{Pagie1997} over the domain $(x,y) \in [-5,5]^2$. 

\begin{align}
  f(x,y) = \frac{1}{1 + x^{-4}} + \frac{1}{1 + y^{-4}} \label{eq:pagie}
\end{align}

We generate $7$ synthetic datasets by uniformly subsampling points from the domain $(x,y) \in [-5,5] \times [-5,5]$. The initial dataset is generated by subsampling a random square grid of dimensions $64 \times 64 = 4096$ points from the domain. To generate the remaining datasets, the length of the grid is iteratively doubled until we subsample a grid of side $4096$ containing over $16$ million points.

\begin{table}[h]
  \caption{Common parameters used for the synthetic dataset benchmarks}
  \begin{center}
    \begin{tabular}[c]{cc}
      \toprule
      \textbf{Parameter} & \textbf{Value} \\
      \midrule
      Runs                      & 10      \\
      Number of Generations     & 50      \\
      Population size           & 50      \\
      Generation Method         & Ramped Half and Half \\
      Fitness Metric            & RMSE    \\
      Crossover probability     & 0.7     \\
      Mutation probability      & 0.25    \\
      Reproduction probability  & 0.05    \\
      Function Set              & $\{+,-,*,\div,\sin ,\cos,\tan\}$ \\
      \bottomrule
    \end{tabular}
    \label{tab:params}
  \end{center}
\end{table}

\Cref{tab:params} lists some common parameters used for training GP models for all libraries. The Ramped Half and Half method was used for tree initialization in all libraries. Root Mean Square Error (RMSE) on the training dataset was chosen as the fitness metric for all population trees. 

\subsubsection{Experimental Results}
\label{sec:results}
A total of $320$ runs were performed across the $5$ different GP libraries to produce the results in this section. We compare the average execution time for $10$ runs for all synthetic datasets, followed by an analysis of the evolution of fitness for \textit{cuml}.

\begin{figure*}[ht]
  \begin{adjustbox}{width=1.2\columnwidth,center}
  \includegraphics{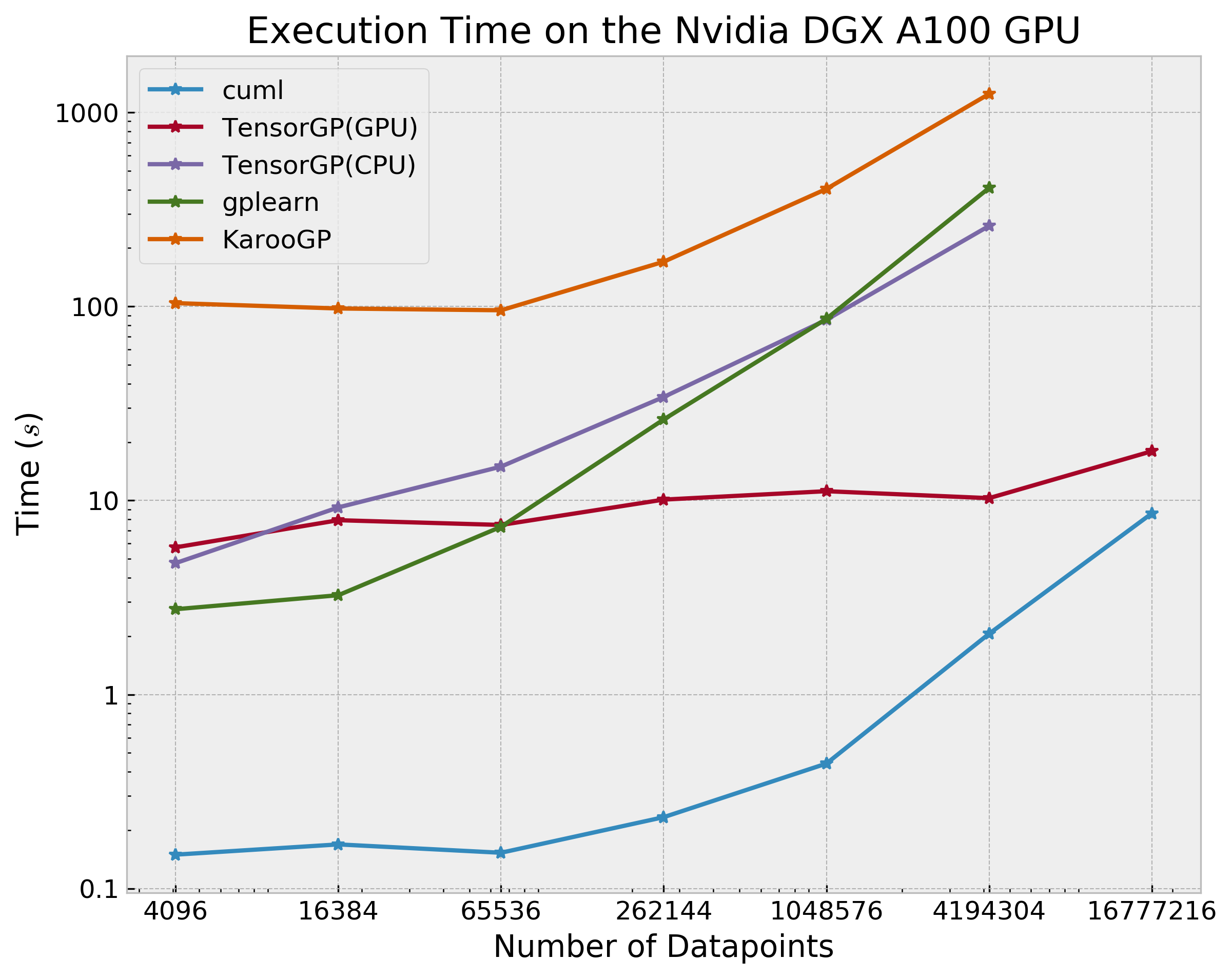}
  \end{adjustbox}
  \caption{Log-Log Plot of Execution Time for various libraries (lower is better). The number of rows considered exponentially increases from $4096$ to over $16$ million.}
  \label{fig:exectimes}
\end{figure*}

\Cref{fig:exectimes} showcases the variation of execution time with increasing dataset size for different datasets, averaged over $10$ runs on each dataset. The same average values are also recorded in \Cref{tab:execavgs}.

\begin{table*}[ht]
  \caption{Table containing mean execution time (lower is better) for a GP run (averaged across $10$ runs) in seconds for different libraries. \textbf{DNF} denotes that the test did not finish within $2$ hours.}
  \begin{adjustbox}{width=1.6\columnwidth,center}
    \begin{tabular}{lrrrrrrr}
    \toprule
      \textbf{\# Rows} &  4096 & 16384 & 65536 & 262144 & 1048576 & 4194304 & 16777216 \\
      \midrule
      \textbf{cuml} &     0.150 &     0.169 &     0.153 &      0.233 &       0.441 &       2.057 &        8.579 \\
            % & STD &    0.024 &     0.015 &     0.005 &      0.008 &       0.033 &       0.022 &        0.020 \\
    %   \midrule
      \textbf{TensorGP (GPU)}  & 5.736 &     7.916 &     7.482 &     10.109 &      11.168 &      10.284 &       17.941 \\
            % & STD &    1.784 &     0.090 &     0.103 &      0.257 &       0.398 &       0.488 &        0.123 \\
    %   \midrule
      \textbf{TensorGP (CPU)} &     4.757 &     9.215 &    14.953 &     34.158 &      85.380 &     260.114 &          \textbf{DNF} \\
            % & STD &    0.272 &     0.631 &     0.047 &      0.106 &       0.503 &       0.336 &          \textbf{DNF} \\
    %   \midrule
      \textbf{gplearn} &     2.753 &     3.250 &     7.320 &     26.167 &      86.369 &     408.217 &          \textbf{DNF} \\
            % & STD &    0.079 &     0.054 &     0.149 &      0.553 &       0.092 &       3.681 &          \textbf{DNF} \\
    %   \midrule
      \textbf{KarooGP} &  104.050 &    97.639 &    95.586 &    170.060 &     403.359 &    1245.424 &          \textbf{DNF} \\
            % & STD &   40.238 &    17.785 &    16.870 &     27.622 &     105.144 &     333.867 &          \textbf{DNF} \\
      \bottomrule
    \end{tabular}
    \end{adjustbox}
    \label{tab:execavgs}
\end{table*}

From \Cref{tab:execavgs}, we note that \textit{cuml} takes $3$ seconds to train the $4$ million row dataset, whereas gplearn takes around $408$ seconds on the same input. When taking a mean average of runtime across all datasets and runs, we achieve an average speedup of $119\times$ in training time with respect to gplearn, with a maximum speedup of $135\times$ in the $4$ million row dataset. 

Since KarooGP uses the TensorFlow \textit{graph} execution model, it is possible that more time is spent in building and modifying the session graph in every generation compared to fitness evaluation. %In comparison, gplearn, uses the \texttt{numpy} library for fitness computations, which can exploit the AVX SIMD instructions on Intel CPUs to achieve greater parallelism. 

% We believe that this job based parallelization again explains why gplearn is faster than TensorGP(CPU) --- which is also optimized to utilize AVX SIMD instructions using TensorFlow's CPU backend.

From the graph in \Cref{fig:exectimes}, as well as \Cref{tab:execavgs}, we see that the TensorGP (GPU) approach consistently outperforms gplearn, KarooGP and TensorGP (CPU) on datasets with more than $65536$ points. Since TensorGP (GPU) uses TensorFlow's GPU backend with eager execution, time is not spent in building an optimized computation DAG for every program. This in turn increases the speed of parallel evaluation.

When comparing \textit{cuml} to the rest of the libraries in \Cref{fig:exectimes} and \Cref{tab:execavgs}, we note that \textit{cuml} is the fastest among all libraries on all inputs. \textit{cuml} also outperforms TensorGP(GPU), since the algorithm batches the computation of fitness across the entire population and dataset. 

\begin{figure}[ht]
  \begin{adjustbox}{width=\columnwidth,center}
    \includegraphics{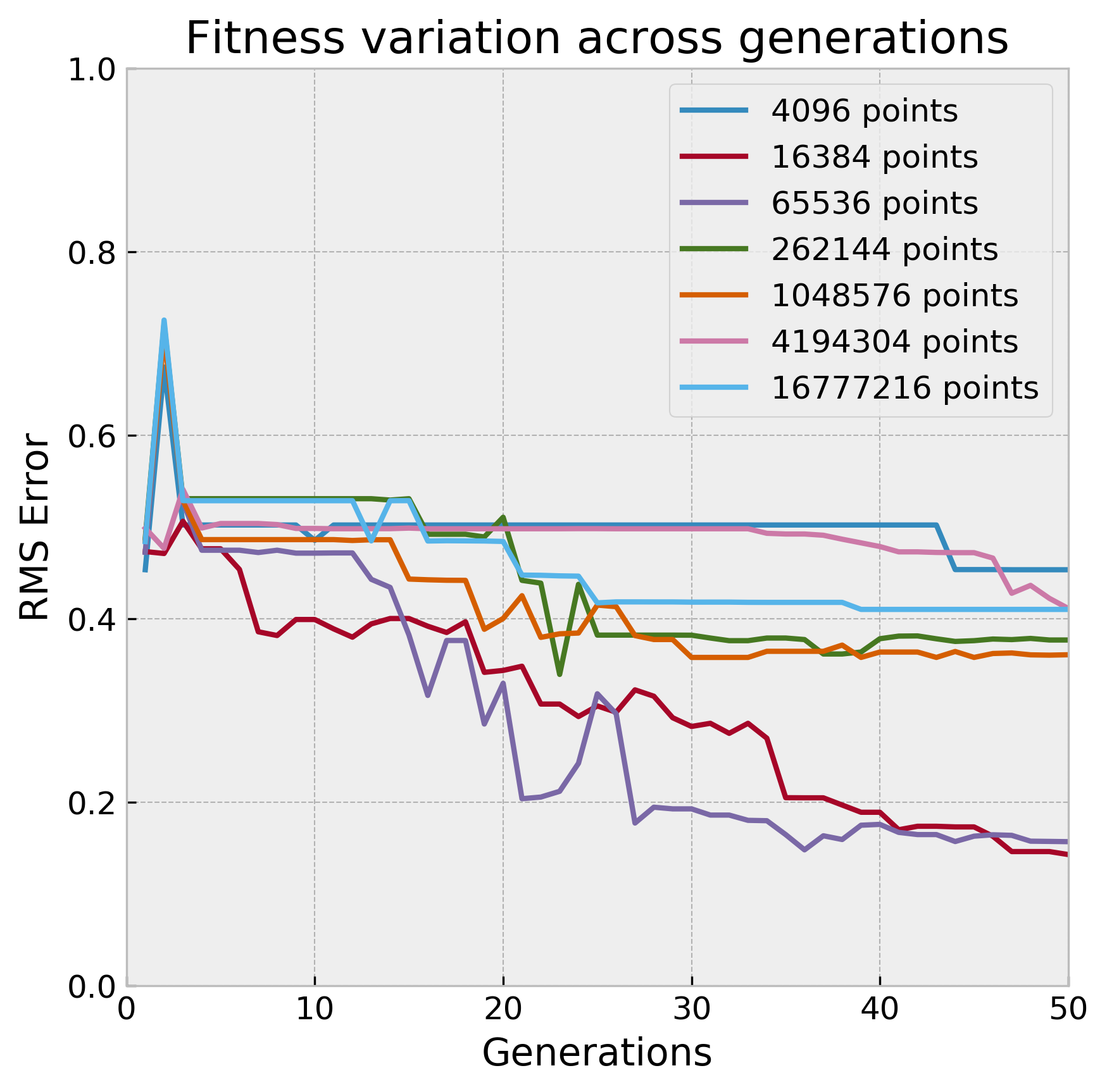}
  \end{adjustbox}
  \caption{Variation of RMS error of with the number of generations for \textit{cuml} on the synthetic datasets.  The error value corresponds to the fitness value of the best tree in every generation.}
  \label{fig:besttrainfit}
\end{figure}

For \textit{cuml}, \Cref{fig:besttrainfit} showcases the variation of the fitness score of the best tree in every generation for all datasets. It is easy to notice that the $1$ million row test set displays a consistently decreasing error with increasing number of generations. However, we notice that the overall fitness value does not decrease even with an increase in the total number of evaluation data points in the same domain. 

Hence, we can conclude that bigger and more granular datasets do not help if we are performing symbolic regression to find the Pagie Polynomial, a result in line with \citep{baeta2021speed}.

\subsection{Large-Scale Benchmarks}
\label{sec:gbm-bench}
We run benchmarks on $6$ large datasets usually used for the comparison of gradient boosting frameworks. In particular, we consider the Airline \citep{airline}, Credit card fraud \citep{fraud}, Higgs \citep{Dua:2019}, and Epsilon \citep{epsilon} datasets for symbolic classification, and the Airline Regression \citep{airline} and YearPredictionMSD datasets \citep{Dua:2019} for symbolic regression. 

\subsubsection{Setup}
\label{subsec:gbm-setup}

\Cref{tab:gbm-datasets} lists all the details of the datasets used in this benchmark. All the classification problems listed are binary classification problems. The train-test split for every dataset is done according to the descriptions provided for every dataset source. 

We run this benchmark only on the \textit{cuml} and \textit{gplearn} library. Since \textit{cuml} is a GPU accelerated re-implementation of \textit{gplearn}, our aim is to achieve similar average test scores for both the libraries along with a decrease in training time.

\begin{table}[ht]
  \caption{All large-scale datasets considered for benchmarks.}
  \begin{adjustbox}{width=\columnwidth,center}
  \begin{tabular}{llll}
    \toprule
    \textbf{Name} & \textbf{Rows} & \textbf{Columns} & \textbf{Task} \\ 
    \midrule
    Airline & 115M & 13 & Classification \\ 
    Airline Regression & 115M & 13 & Regression \\ 
    Fraud & 285K & 28 & Classification \\ 
    Higgs & 11M & 28 & Classification \\ 
    Year & 515K & 90 & Regression \\ 
    Epsilon & 500K & 2000 & Classification \\ 
    \bottomrule
    \end{tabular}
  \end{adjustbox}
  \label{tab:gbm-datasets}
\end{table}

To speed up the execution times for gplearn, every GP run was parallelized using $8$ jobs(using the Python \texttt{joblib} library).

\Cref{tab:gbm-params} lists all common parameters used for training GP models for both \textit{cuml} and \textit{gplearn}. RMS Error is again used as the fitness metric for all regression datasets, while Logistic loss is used as the fitness metric for all the binary classification datasets. 

\begin{table}[ht]
  \caption{Common parameters used for the large dataset benchmarks.}
  \begin{center}
    \begin{tabular}[c]{cc}
      \toprule
      \textbf{Parameter} & \textbf{Value} \\
      \midrule
      Runs                      & 3      \\
      Number of Generations     & 50      \\
      Population size           & 35      \\
      Tournament size           & 4       \\ 
      Generation Method         & Ramped Half and Half \\
      Fitness Metric            & RMSE / Logistic Loss    \\
      Crossover probability     & 0.7       \\
      Subtree mutation probability & 0.1    \\
      Point mutation probability & 0.1      \\
      Hoist mutation probability & 0.05     \\
      Reproduction probability  & 0.05      \\
      Parsimony coefficient     & 0.01      \\
      Function Set              & $\{+,-,*,\div,\sin ,\cos,\tan\}$ \\
      \bottomrule
    \end{tabular}
    \label{tab:gbm-params}
  \end{center}
\end{table}

\subsubsection{Experimental Results}
\label{subsec:gbm-results}
A total $36$ GP runs were performed on the \textit{cuml} and \textit{gplearn} to produce the results in this section. We compare the average training time taken by both libraries, followed by a comparison of results on the test dataset for both libraries. 

\begin{figure*}[ht]
  \begin{adjustbox}{width=1.5\columnwidth,center}
  \includegraphics{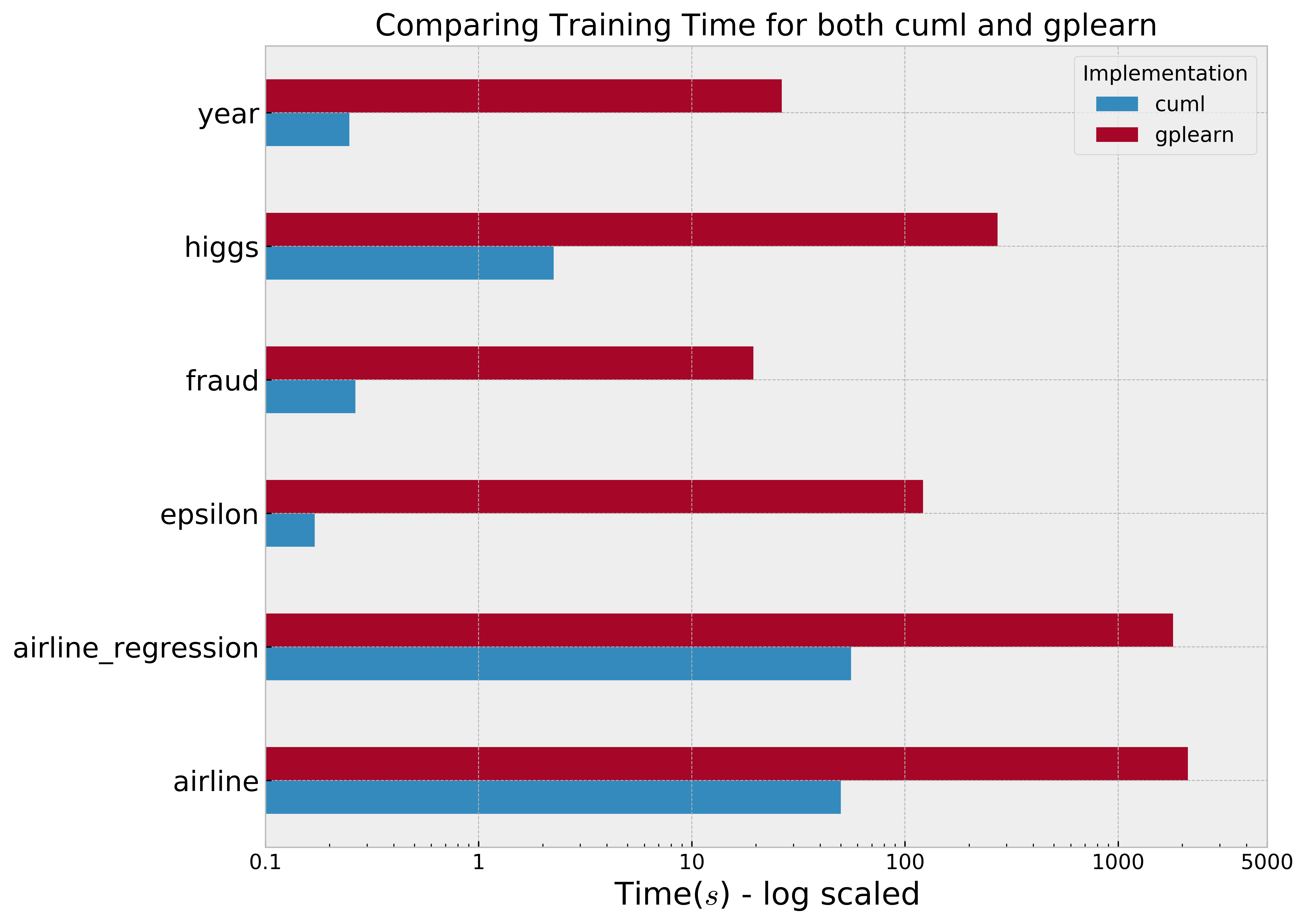}
  \end{adjustbox}
  \caption{Bar graph showcasing the difference in training time for both \textit{cuml} and \textit{gplearn} (lower is better). A speedup in training time is observed for all datasets when using \textit{cuml}. }
  \label{fig:gbm-exectimes}
\end{figure*}

\Cref{fig:gbm-exectimes} showcases the variation of execution time for both \textit{cuml} and \textit{gplearn} for all the datasets listed in \Cref{tab:gbm-datasets}. The same values can be found in \Cref{tab:gbm-exectimes}. 

From \Cref{tab:gbm-exectimes} we find that \textit{cuml} achieves a mean speedup of $40\times$ in training with respect to \textit{gplearn} across all the datasets, with a maximum speedup of $713\times$ in the Epsilon dataset. The tremendous speedup observed in the Epsilon dataset (which has $2000$ columns) can be attributed to the coalesced memory access resulting from the column-major order storage of the input data in GPU memory. On the Airline and Airline regression datasets, $2$ datasets with more than $100$ million rows, \textit{cuml} achieves an average speedup of $42\times$ and $32\times$ with respect to \textit{gplearn}(parallelized using $8$ jobs) in training time.

\begin{table}[ht]
  \caption{Table containing mean execution time (lower is better) for a GP run (averaged across $3$ runs) in seconds for both \textit{cuml} and \textit{gplearn}.}
  \begin{adjustbox}{width=0.75\columnwidth,center}
  \begin{tabular}{lrr}
        \toprule
        % {}               & \textbf{Time(s)}     & {} \\
        \textbf{Dataset} &   \textbf{cuml} &  \textbf{gplearn} \\
        \midrule
        Airline            & 50.088 & 2122.619 \\
        Airline Regression & 55.935 & 1810.712 \\
        Epsilon            &  0.170 &  121.543 \\
        Fraud              &  0.264 &   19.448 \\
        Higgs              &  2.250 &  272.140 \\
        Year               &  0.248 &   26.444 \\
        \bottomrule
        \end{tabular}
  \end{adjustbox}
  \label{tab:gbm-exectimes}
\end{table}

\Cref{tab:gbm-clf} compares the symbolic classification accuracy on the test split of all classification datasets for both \textit{cuml} and \textit{gplearn} on the best program at the end of the GP run. We note that for all the datasets, the test accuracy for both the libraries lie with $5\%$ of each other, with \textit{cuml} outperforming \textit{gplearn} in the Higgs and  Epsilon datasets.

\begin{table}[ht]
  \caption{Table containing classification accuracy values (normalized to $[0,1]$) on the test split of all datasets for both \textit{cuml} and \textit{gplearn} (higher is better).}
  \begin{adjustbox}{width=0.75\columnwidth,center}
    \begin{tabular}{lrr}
    \toprule
    \textbf{Classification} &      \textbf{cuml} &   \textbf{gplearn} \\
    \textbf{Datasets} &     {}      &      {}     \\
    \midrule
    Airline &  0.474572 &  0.525428 \\
    Epsilon &  0.560250 &  0.520487 \\
    Fraud   &  0.998280 &  0.998280 \\
    Higgs   &  0.567743 &  0.530103 \\
    \bottomrule
    \end{tabular}
  \end{adjustbox}
  \label{tab:gbm-clf}
\end{table}

\Cref{tab:gbm-reg} compares the test RMS Error obtained at the end of the GP run for both \textit{gplearn} and \textit{cuml}. We again notice that the error values obtained for both libraries are almost the same, especially in the case of the Airline Regression dataset. For the Year prediction dataset, we see that \textit{cuml} slightly outperforms the \textit{gplearn} with respect to the test RMS error. 

\begin{table}[ht]
  \caption{Table containing RMS Error values on the test split of all datasets for both \textit{cuml} and \textit{gplearn} (lower is better).}
  \begin{adjustbox}{width=0.75\columnwidth,center}
  \begin{tabular}{lrr}
    \toprule
    \textbf{Regression} &    \textbf{cuml} &  \textbf{gplearn} \\
    \textbf{Datasets}            &   {}      &  {}        \\
    \midrule
    Airline Regression &  31.024 &   30.900 \\
    Year               & 825.138 &  840.209 \\
    \bottomrule
    \end{tabular}
  \end{adjustbox}
  \label{tab:gbm-reg}
\end{table}

We note that \Cref{tab:gbm-clf,tab:gbm-reg} do not allow to draw a conclusion on which library produces better results for symbolic regression in general. This is because the final fitness values achieved are highly sensitive to the initialization of programs, which is randomized. Rather, the only goal of the test metrics is to help us verify similar behaviour of fitness values across libraries on the same datasets, when initialized with the same training hyperparameters.

% In the next section, we shall list out some conclusions and possible optimizations and feature additions to the current implementation of \textit{cuml}. 
\section{SUMMARY AND FUTURE WORK}
\label{chap:conclusion}

In this paper, we accelerated genetic programming based symbolic regression and classification on the GPU. We do this by parallelizing the selection and evaluation step of the generational GP algorithm. We introduce a prefix-list representation for expression trees, which are then evaluated using an optimized stack on the GPU. Fully vectorized routines for standard loss functions are also provided, which can be used as fitness functions during the training of a genetic population of programs. At the end of a run, our algorithm returns the entire set of evolved programs for all generations. The most optimal program from the last generation of programs can be used as a potential symbolic regressor or classifier. 

Our experimental results using both synthetic and large scale datasets indicate that our implementation of genetic programming is significantly faster when compared to existing CPU and GPU parallelized frameworks. Performance benchmarks indicate that our implementation achieves an average speedup $119\times$ and $40\times$ against \texttt{gplearn}, a CPU parallelized GP library on the synthetic and large scale datasets respectively.

In the future, we are planning to add support for custom function sets and loss functions during GP runs. We are also planning to address the problem of multiple memory transfers between the CPU and the GPU by slightly modifying the underlying program representation without compromising on the stack based evaluation model for programs. In addition, we are also planning to implement a Python layer over the current CUDA/C++ based implementation of the algorithm.
% with a \texttt{scikit-learn}-like API 
% is also planned to be implemented.

% \section*{Acknowledgement}
% This paper and the research behind this would not have been possible without the generous support from NVIDIA and IIT Madras, and the resulting collaboration facilitated by the NVIDIA AI Technology Center (NVAITC) program.

This paper and the research behind this would not have been possible without the generous support of the NVIDIA AI Technology Center (NVAITC), which facilitated this collaborative work between NVIDIA and IIT Madras.

\bibliographystyle{IEEEtran}
\bibliography{references}
\end{document}